\documentclass[runningheads]{llncs}
\usepackage{graphicx}
\usepackage{amsmath}
\usepackage[boxed,ruled,vlined,linesnumbered,scleft]{algorithm2e}
\usepackage{hyperref}
\usepackage{xcolor}

\begin{document}
\title{General Boolean Formula Minimization\\ with QBF Solvers\thanks{This project has received funding from the European Union’s Horizon 2020 research and innovation program under the Marie Skłodowska-Curie grant agreement No.~860621 and the MICINN project PROOFS
(PID2019-109137GB-C21).}}
%
%
\author{Eduardo Calò\inst{1}\orcidID{0000-0003-3881-8994} 
\and Jordi Levy\inst{2}\orcidID{0000-0001-5883-5746}
}
\authorrunning{E. Calò and J. Levy}
%
\institute{Utrecht University, the Netherlands\\
\email{e.calo@uu.nl}\\
\url{https://www.uu.nl/staff/ECalo} 
\and IIIA, CSIC, Spain\\
\email{levy@iiia.csic.es}\\
\url{https://www.iiia.csic.es/~levy}}
\maketitle              
\begin{abstract}
The minimization of propositional formulae is a classical problem in logic, whose first algorithms date back at least to the 1950s with the works of Quine and Karnaugh. Most previous work in the area has focused on obtaining minimal, or quasi-minimal, formulae in conjunctive normal form (CNF) or disjunctive normal form (DNF), with applications in hardware design. In this paper, we are interested in the problem of obtaining an equivalent formula in any format, also allowing connectives that are not present in the original formula. We are primarily motivated in applying minimization algorithms to generate natural language translations of the original formula, where using shorter equivalents as input may result in better translations. Recently, Buchfuhrer and Umans have proved that the (decisional version of the) problem is $\Sigma_2^p$-complete. 

We analyze three possible (practical) approaches to solving the problem. First, using brute force, generating all possible formulae in increasing size and checking if they are equivalent to the original formula by testing all possible variable assignments. Second, generating the Tseitin coding of all the formulae and checking equivalence with the original using a SAT solver. Third, encoding the problem as a Quantified Boolean Formula (QBF), and using a QBF solver. Our results show that the QBF approach largely outperforms the other two.

\end{abstract}
\section{Introduction}\label{sec:intro}

The minimization of complex Boolean expressions is a longstanding problem in logic. The first algorithms developed in the 1950s, e.g., the works of Quine, McCluskey \cite{quine1,quine2,mcclus}, and Karnaugh \cite{Karnaugh} paved the way for extensions and optimizations in the following years (e.g., the Petrick's method \cite{petrick1956direct}, and the Espresso heuristic logic minimizer \cite{brayton1982comparison}, \emph{i.a.}). These works have focused on obtaining minimal equivalent representations in specific canonical forms (e.g., conjunctive normal form (CNF) or disjunctive normal form (DNF)), and confined the studies to a limited set of connectives. Here, we are interested in the \emph{general} Boolean formula minimization, where no assumptions are made in the form of the input formula or the output. In fact, our minimization methods allow us to use distinct sets of connectives for the input and the output.

We frame Boolean minimization (i.e., finding the logically equivalent shortest formula(e) to a given one) as a Quantified Boolean Formulae (QBF) satisfiability problem and design an algorithm that consistently finds the shortest equivalents of a given formula. We compare this algorithm with a brute force baseline, and an approach based on SAT.

\subsubsection*{Motivation.}
We have two distinct motivations behind our work, which lay very far from each other. Our first motivation is shared in~\cite{qbf2epr}, where the authors present {\tt qbf2epr}, a tool that translates QBF to formulas in effective propositional logic (EPR). Their aim is to generate benchmarks for EPR and compare solvers for QBF and EPR. Similarly, our formula minimization problem, encoded as QBF, generates benchmarks for QBF solvers and allows us to compare SAT and QBF techniques. The automated deduction community is divided into sub-communities (e.g., SAT, QBF, SMT, MaxSAT, EPR), which try to solve distinct classes of problems, from SAT which is NP-complete, to EPR which is NEXPTIME-complete, passing by QBF which is PSPACE-complete, and each one has its own competition and set of benchmarks. However, many ideas that proved effective in one area (like learning in SAT) have been exported to others. In this sense, problems that could be solved with two distinct technologies, like ours, contribute to comparing the level of maturity reached in each area.

Our second motivation relates to a use case of minimization algorithms in natural language processing. Grasping the meaning of logical formalisms is a crucial task for many scholars, yet sometimes even experienced logicians might have trouble deciphering a complex formula. Techniques from natural language generation \cite{reiter_dale_2000,gatt2018survey}, and in particular logic-to-text generation methodologies \cite{ranta2011transl-lang-logic,calo-etal-2022-enhancing}, can be used for simplifying and translating logical formulae into optimally intelligible text in natural languages (NLs) (such as English, Mandarin, or Korean), which can effectively explain formulae to systems' users. For example, given the following first-order formula:
\[
\exists x ( Problem ( x ) \wedge \forall y ( Researcher ( y ) \rightarrow Interested ( y , x ) ) )
\]

\noindent we want a system that can automatically generate a faithful and comprehensible explanation, via the following (or another semantically equivalent) text:

\vspace{0.3cm}
\centerline{\textit{There is a problem that every researcher finds interesting.}}
\vspace{0.3cm}

What are the characteristics that a formula should have to become a suitable input for a logic-to-text translation system? One aspect that one might want to look at is length. Brevity has surrounded linguistic debate at least since \cite{grice1975logic}. Arguably, shorter utterances should be preferred over longer ones and unnecessary prolixity should be avoided. This principle might also apply to logical formulae. Intuitively, a short formula, rather than a longer logical equivalent, should be better suited to be translated into NL. In this paper, we tackle exclusively the logical aspect of the problem. We focus on propositional logic, a formalism in which equivalence is decidable, and limit our examination to formulae's length,\footnote{We define length as the number of symbols (i.e., predicates and connectives, parentheses excluded) contained in a formula.} aiming only for the shortest equivalents to a given formula.

\subsubsection*{Related Work: Formula Minimization.} \label{sec:minim}
Boolean formula minimization is a natural optimization problem in the second level of the Polynomial-Time Hierarchy $\Sigma^p_2$. Indeed, the problem is used by \cite{garey1979computers} to motivate the definition of the Polynomial Hierarchy. Its decisional version can be formulated as the following problem: Given a Boolean formula, prove the existence of a (smaller) formula (in the same set of variables) that gets the same evaluation of the given formula, for all possible assignments of the variables. The fact that both sets of quantified variables, as well as the time to evaluate the formulae, are bounded in the input proves its inclusion in $\Sigma^p_2$. As we will see in Section~\ref{sec:algorithms}, this corresponds to our brute-force algorithm. It is assumed that both the given formula and its minimization are circuits or formulae of the same form. However, in our implementation, we leave open the possibility to use distinct sets of connectives. Apart from some completeness proofs for some particular forms of the input and output, the proof for the general form had eluded researchers until \cite{complexityminimization} proved $\Sigma^p_2$-completeness of the problem.

The optimization of complex Boolean expressions has been studied extensively in electronic circuits, where practical matters (i.e., complex circuits take up physical space and costs more resources in their implementation) make it crucial to find optimal circuit representations. Well-known minimization methods include the Quine-McCluskey algorithm \cite{quine1,quine2,mcclus} and the Karnaugh map \cite{Karnaugh}. In the Karnaugh map, Boolean results are transferred from a truth table onto a two-dimensional grid, where each cell position represents one combination of input conditions, while each cell value is the corresponding output value. Optimal groups of $0$s and $1$s are identified, which represent the terms of a canonical form that can be used to write a minimal expression. The Quine–McCluskey algorithm finds all the prime implicants of a function and uses them in a chart to find (i) the essential prime implicants of the function, and (ii) other prime implicants that are necessary to cover the function. The method is functionally identical to the Karnaugh map, but its tabular form makes it more efficient to employ in computer systems.

However, despite this long history of research and attempts to extend well-established methods (e.g., \cite{turton_extending_1996} tries to implement the {\tt XOr} operator in the Quine-McCluskey algorithm), most work has focused on a limited set of connectives and canonical forms (e.g., CNF or DNF). For our scope, we need a more general approach where all connectives could be tackled, on demand.

\subsubsection*{Quantified Boolean Formulae.} 

Quantified Boolean Formulae (QBFs) are an extension of propositional logic, where universal and existential quantifications are allowed \cite{BeyersdorffJLS21}. The use of quantifiers results in a greater expressive power than classic propositional logic. If all variables occurring in a QBF $\phi$ are bound, then $\phi$ is called \textit{closed}. QBFs often assume a canonical prenex conjunctive normal form (PCNF) $\phi = \exists \vec{x} \forall \vec{y} \exists \vec{z}\cdots \psi$, where the portion containing only quantifiers and bound variables is called the \textit{prefix}, followed by $\psi$ that is a quantifier-free Boolean formula with conjunctions over clauses, called the \textit{matrix}.

The QBF satisfiability problem \cite{giunchiglia2009reasoning} consists of determining, for a given QBF $\phi$, the existence of an assignment for the free variables, such that $\phi$ evaluates to true under this assignment. Hence, $\phi$ is true iff, there exists a truth assignment to $\vec{x}$, such that, for all truth assignments to $\vec{y}$, there exists a truth assignment to $\vec{z}$,\dots such that $\psi$ is true. Several QBF solvers have been developed over time,\footnote{\url{http://www.qbflib.org}} and applications of QBFs technologies range from AI to planning \cite{cashmore2010planning,diptarama2016qbf,shukla_survey_2019}. QBF solvers only use to provide the instantiation of most externally existentially-quantified variables $\vec{x}$, since for the other ones, instantiation depends on previous universal variables $\vec{z}= f(\vec{y})$. In this work, we exploit QBFs to encode and solve the Boolean minimization problem.

\subsubsection{Structure of the Paper.}
The rest of the paper is structured as follows. 
Section~\ref{sec:algorithms} introduces the algorithms that we employ in our experiments and the QBF encoding we develop. Section~\ref{sec:experiments} illustrates the experiments we carry out, comparing the three aforementioned approaches, and shows the results. We present some reflections on possible future directions in Section~\ref{sec:conclusions}.

\section{Algorithms}\label{sec:algorithms}

In our experimentation, we analyze three algorithms that we will call brute-force, SAT-based, and QBF-based.

\begin{algorithm}[t]
\KwIn{$\phi$} 
\KwOut{a minimal equivalent formula $\psi$} 
\DontPrintSemicolon
\SetKwFunction{FMain}{main}
\SetKwFunction{FEquivalent}{equivalent}
\SetKwProg{Fn}{Function}{:}{}
 \Fn{\FEquivalent{$\phi,\psi$}}{
    \ForEach{Assignment $I:Var(\phi)\to\{0,1\}$}{
        \If{$I(\phi)\neq I(\psi)$}{
            \KwRet{$true$}
        }
    }
\KwRet{$false$}
}
\Fn{\FMain{$\phi$}}{
\ForEach{$i=1,\dots,|\phi|$}{
    \ForEach{formula s.t. $|\psi|=i$ and $Var(\psi)\subseteq Var(\phi)$}{
        \If{$equivalent(\phi,\psi)$}{
            \KwRet{$\psi$}
        }
  }
}
}
\caption{Brute-force algorithm}\label{alg:brute-force}
\end{algorithm}

The brute-force algorithm (see Alg.~\ref{alg:brute-force}) is the algorithm that we mention in Section~\ref{sec:minim} as proof that formula minimization is in $\Sigma_2^p$. Two formulae $\phi$ and $\psi$ are equivalent iff $\phi\leftrightarrow \psi$ is a tautology. Like TAUT, the formula equivalence problem is CoNP-complete. However, considering that we test the equivalence for all formulae $\psi$ smaller than $\phi$, the average time for the calls to {\tt equivalent($\phi,\psi$)} is the same as considering $\psi$ a random formula smaller than $\phi$. Then, the \emph{average} time required by the function call is only $\mathcal{O}(|\phi|)$. Notice that in this situation, half of the calls finish after checking one assignment, $1/4$ after checking two assignments, etc. Hence, on average we check $\sum_{i=1}^{2^{|\phi|}} i\frac{1}{2^i}< 2$ assignments in every call.

We can also estimate the number of calls to this function as follows. The number of distinct complete trees of size $n$ that we can construct with $k$ binary symbols is $k^n$. If the trees can have any form, then the computation is more complicated. Let $\mathcal{C}$ be the set of possible binary symbols (hence, we are not considering {\tt Not}) and $\mathcal{V}$ be the set of possible leaves. The number of forms of trees with $m$ binary nodes and $m+1$ leaves is given by the recurrence $f(m) = \sum_{i=0}^{m-1} f(i)\,f(m-i-1)$ that define the Catalan numbers $C_m$. The number of distinct trees will be $C_m\,|\mathcal{C}|^m\,|\mathcal{V}|^{m+1}$. Using this Stirling approximation, this can be approximated as $\frac{4^m}{\sqrt{\pi}m^{3/2}}\,|\mathcal{C}|^m\,|\mathcal{V}|^{m+1}$. As a function of the tree size $n=2m+1$, this is $\mathcal{O}((4|\mathcal{C}||\mathcal{V}|)^{n/2} / n^{3/2})$ calls to the {\tt equivalent} function.

\begin{algorithm}[ht]
\KwIn{$\phi$} 
\KwOut{a minimal equivalent formula $\psi$} 
\DontPrintSemicolon
\SetKwFunction{FMain}{main}
\SetKwFunction{FEquivalent}{equivalent}
\SetKwFunction{FTseitin}{tseitin}
\SetKwProg{Fn}{Function}{:}{}

 \Fn{\FTseitin{$\phi,x$}}{
 \If{$\phi = \phi_1 \wedge \phi_2$}{
 $y_1, y_2 := freshvars()$\;
 \KwRet{$tseitin(\phi_1,y_1)\cup tseitin(\phi_2,y_2)\cup
 CNF(\{x \leftrightarrow y_1\wedge y_2\})$}
 }
 $\cdots$ /* Similarly for other connectives or variables */
}
 \Fn{\FEquivalent{$\phi,\psi$}}{
        $x_1,x_2 := freshvars()$\;
        $\Gamma := tseitin(\phi,x_1)\cup tseitin(\psi,x_2)\cup CNF(\{\neg(x_1\leftrightarrow x_2)\})$\;
        \KwRet{$SAT(\Gamma)\neq satisfiable$}
}
\Fn{\FMain{$\phi$}}{
\ForEach{$i=1,\dots,|\phi|$}{
    \ForEach{formula s.t. $|\psi|=i$ and $Var(\psi)\subseteq Var(\phi)$}{
        \If{$equivalent(\phi,\psi)$}{
            \KwRet{$\psi$}
        }
  }
}
}
\caption{SAT-based algorithm}\label{alg:SAT-based}
\end{algorithm}

The second algorithm (see Alg.~\ref{alg:SAT-based}) is based on the use of a SAT solver and the Tseitin encoding of the two formulae that we want to prove equivalent. Given two formulae $\phi$, $\psi$, we can find, in linear time $|\phi|+|\psi|$, a CNF formula $\Gamma$ such that the two formulae are equivalent iff $\Gamma$ \emph{is not} satisfiable. At first sight, it may not look reasonable to use a SAT solver to check a property that on average only requires linear time. However, experiments show that, in practice, we still can get some gain with respect to the brute-force algorithm (see Section~\ref{sec:experiments}).

\begin{algorithm}[ht]
\KwIn{$\phi$} 
\KwOut{a minimal equivalent formula $\psi$} 
\DontPrintSemicolon
\SetKwFunction{FMain}{main}
\SetKwFunction{FEquivalent}{equivalent}
\SetKwFunction{FScheme}{scheme}
\SetKwProg{Fn}{Function}{:}{}

\Fn{\FScheme{$\delta,z$}}{
    $x_{false} := freshvar(\exists^{(1)})$\;
    $\Gamma := CNF(\{x_{false} \rightarrow \neg z\})$\;
    \ForEach{$y \in \forall^{(1)}$}{
        $x_{y} := freshvar(\exists^{(1)})$\;
        $\Gamma := \Gamma \cup CNF(\{x_y \rightarrow(z\leftrightarrow y\})$
    }
    \If{$\delta>0$}{
        $z_1,z_2 := freshvar(\exists^{(2)})$\;
        $\Gamma := \Gamma\cup scheme(\delta-1,z_1) \cup scheme(\delta-1,z_2)$\;
        \ForEach{$c\in \mathcal{C}$}{
            $x_c := freshvar(\exists^{(1)})$\;
            $\Gamma := \Gamma\cup CNF(\{x_c\rightarrow(z\leftrightarrow z_1\ c\ z_2)\})$
        }
    }
    \KwRet{$\Gamma\cup CNF(\{x_{false} + \sum_{y\in\forall^{(1)}} x_y + \sum_{c\in \mathcal{C}} x_c = 1\})$}
}
\Fn{\FEquivalent{$\phi,\delta$}}{
    $z_1,z_2 := freshvars(\exists^{(2)})$\;
    $\forall^{(1)} := Vars(\phi)$\;
    $\Gamma := tseitin(\phi,z_1)\cup scheme(\delta,z_2)\cup CNF(\{z_1\leftrightarrow z_2\})$\;
    \KwRet{$QBF(\Gamma)=true$}
}
\Fn{\FMain{$\phi$}}{
\ForEach{$\delta=1,\dots,depth(\phi)$}{
    \ForEach{formula s.t. $depth(\psi)=i$ and $Var(\psi)\subseteq Var(\phi)$}{
        \If{$equivalent(\phi, \delta)$}{
            \KwRet{$\psi$}
            }
        }
    }
}
\caption{QBF-based algorithm}\label{alg:QBF-based}
\end{algorithm}

The third algorithm (see Alg.~\ref{alg:QBF-based}) is based on the use of a QBF solver. Here, instead of testing every possible minimal formula $\psi$, we test every possible depth $\delta$. This supposes a significant improvement since there is a linear number of depths to try, instead of an exponential number of formula candidates. Second, instead of a Tseitin encoding of the candidate, we compute a \emph{scheme of the candidate}. The equivalence between the original formula and this scheme can be encoded as a QBF formula with three quantifier alternations: $\exists \vec{x}.\forall \vec{y}.\exists \vec{z}.\Gamma$. In Alg.~\ref{alg:QBF-based}, these three sets of variables are represented as $\exists^{(1)}$, $\forall^{(1)}$, and $\exists^{(2)}$ and individual variables are named $x$, $y$, and $z$, respectively. If the QBF formula is true, the values we got for variables $x\in\exists^{(1)}$ will encode the minimal formula. Notice that QBF solvers only provide the instantiations of the most external existentially-quantified variables, since the values of the other existentially-quantified variables depend on more externally universally-quantified variables.\footnote{In the case of using a QBF solver unable to provide these instantiations, we cannot compute the minimal equivalent formula.} Basically, for every node $i$ of the scheme and every truth constant, variable $y$ of the original formula, or connective $c$, we have a variable $x_c^i$ that gets the value true when at position $i$ we have the connective $c$ (resp. variable $y$ or constant). The constraints $CNF(\{x_{false} + \sum_{y\in\forall^{(1)}} x_y + \sum_{c\in \mathcal{C}} x_c = 1\})$ ensure that one, and only one, of them get the value true. Variables in $\forall^{(1)}$ are just the set of variables in the original formula. The original formula and the scheme are equivalent if \emph{for all} assignments to these variables, both the original formula and scheme get the same evaluation. Variables in $z^i\in\exists^{(2)}$ encode the truth values for every possible subformula at position $i$ of the scheme or of the Tseitin encoding of the original formula. The clauses in the QBF formula encode restrictions of the form 
\[
\begin{array}{l}
x^i_{\mbox{\scriptsize false}}\to \neg z^i\\[1mm]
x^i_y \rightarrow(z^i\leftrightarrow y)\\[1mm]
x_c^i \rightarrow (z^i \leftrightarrow z^{i\cdot 1}\, c\, z^{i\cdot 2})
\end{array}
\]
The intended meanings of these constraints are: if at position $i$ of the scheme we have the constant false, the sub-scheme is evaluated to false, if there is an original variable $y\in\forall^{(1)}$, it is evaluated to $y$, and if there is a connective $c\in \mathcal{C}$, then the sub-scheme gets the same value as the connective $c$ operated on the evaluations $z^{i\cdot 1}$ of the left-child of $i$ and the evaluation $z^{i\cdot 2}$ of the right-child.

Notice that the size of the QBF formula we get is $\mathcal{O}(2^{depth(\phi)}\cdot (|\mathcal{C}|+|Vars(\phi)|))$. Assuming that the original formula is balanced, in practice $2^{depth(\phi)} \approx |\phi|$. Therefore, we could consider it a polynomial encoding. Notice also that we do not make a profit from the commutativity and associativity of most connectives.

When in a node of the scheme we put a {\tt Not}, we only use one of the children, and in the case of putting a variable, we do not use any of the children. To avoid useless search in the QBF solver we can force all these useless nodes to be fixed to a dummy value by adding the constraints $CNF(\{x^i_{\neg}\to x^{i\cdot 2}_{dummy}\})$ and $CNF(\{(x^i_y \vee x^i_{dummy} \vee x^i_{false}) \to (x^{i\cdot 1}_{dummy} \wedge x^{i\cdot 2}_{dummy})\})$.

In this approach, we only bound the depth of the scheme. If we also want to limit its size, we can add the encoding of some cardinality constraint that bound the number of nodes in the schema that are distinct from the dummy: 
$\sum_i \neg x^i_{dummy} \leq size\_bound$

\section{Experiments and Results}\label{sec:experiments}

We conduct some experimentation with our algorithms. The three algorithms are implemented in Python 3 and are publicly available at \url{https://gitlab.nl4xai.eu/eduardo.calo/QBF-boolean-minimization}. In the case of the SAT-based algorithm, we use the Python module {\tt python-sat}\footnote{\url{https://github.com/pysathq/pysat}} as SAT solver. In the case of the QBF-based algorithm, we use the {\tt CAQE}~\cite{CAQE,CAQE2} QBF solver, although any other QBF solver that accepts QDIMACS standard\footnote{\url{http://www.qbflib.org/qdimacs.html}} input and output may be used. 

For every size in $s=1,\dots,20$, we generate $100$ random formulae of size $s$ and minimize them using the three algorithms. We make sure that all syntactically distinct formulae are generated with the same probability. However, we do not take into account the commutativity and associativity of connectives or other formula equivalences. Formulae of size $s$ are generated over a set of $\sqrt{s}$ variables\footnote{Using $s/c$ or $s^c$ variables does not seem to affect substantially the results.} and connectives $\tt\mathcal{C}=\{Not, And, Or\}$ and minimization are searched among formulae with connectives $\tt\mathcal{C}'=\{Not, And, Or, Implies\}$.

We use a cluster with $11$ calculating nodes with $2$ Intel Xeon CPUs at $2.2 GHz$ with $10$ cores/CPU and $92 GB$ of RAM. We set a time-out of $20,000s$. The brute-force and the SAT-based algorithms reach the time-out in some instances for $s=15,18,19,20$. These values are not considered in the computation of the mean and median times. Therefore, these mean and median values are abnormally low. 

\begin{figure}[t]
\hspace{-2mm}
\includegraphics[width=0.52\textwidth]{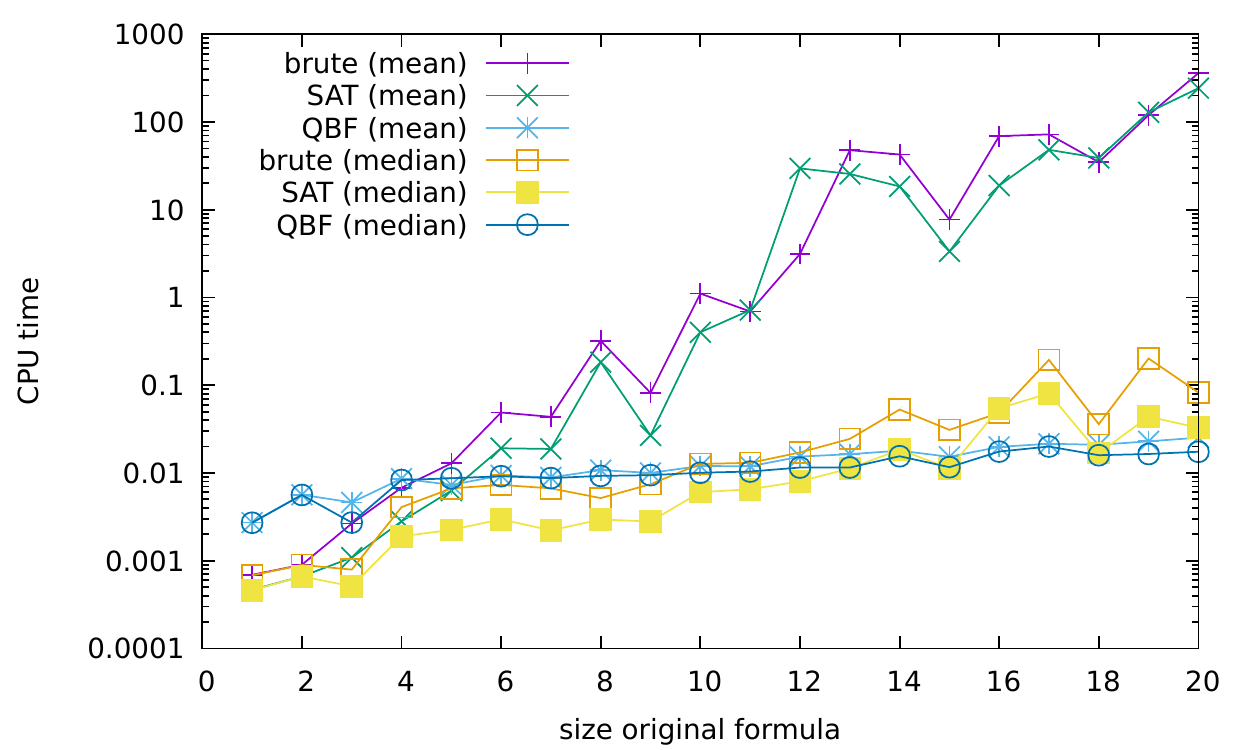}
\hspace{-20mm}\hfill
\includegraphics[width=0.52\textwidth]{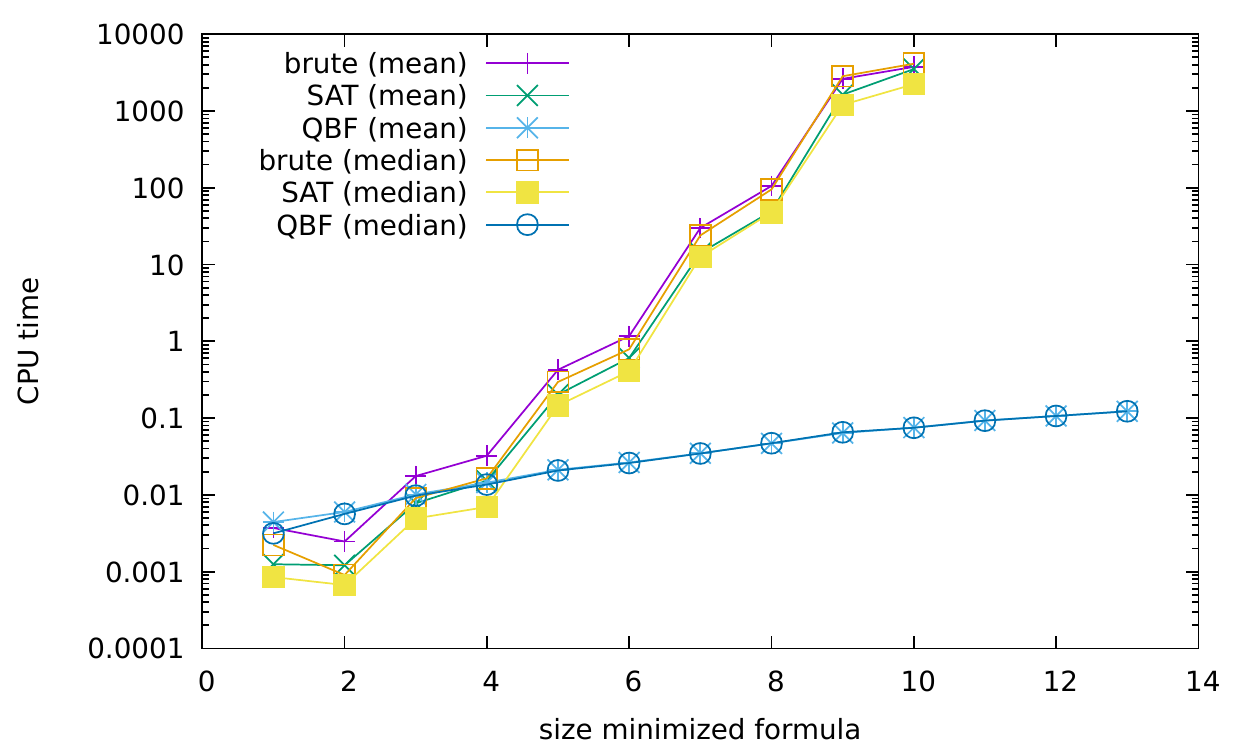}
\hspace{-4mm}
\caption{Average and median time required by the three algorithms with respect to the size of the original formula (left) and the resulting minimized formula (right).}\label{fig:times}
\end{figure}

In Figure~\ref{fig:times} (left), we show the average and median (logarithm of) CPU time required by each one of the algorithms as a function of the size of the input formula. We clearly observe that the QBF-based algorithm outperforms the other two algorithms, which seem to require exponential time on the size of the input. We also observe that the SAT-based is consistently better than the brute-force algorithm (a constant distance between the functions, in logarithmic axes, means an improvement of constant factor). This is quite surprising since, as we mention in Section~\ref{sec:algorithms}, the computation of the formula equivalence can be done in linear average time. It is also remarkable that, in the case of brute-force and SAT-based, there is a significant difference between the average and median time. The reason, as we comment in detail below, is the significant variability in the times required by each instance. The same effect produces a fluctuation in the values of the average time. We can conclude that, although in most of the instances (attending to the median), the three algorithms minimize the formula in less than one second, for sizes smaller than $20$, just a few instances make brute-force and SAT-based require around $1h$ on average when the size is around~$20$.

In Figure~\ref{fig:times} (right), we show the average and median (logarithm of) CPU time as a function of the size of the obtained minimal formula. Here the differences between the mean and median times are smaller. Hence, we can conclude that the size of the output determines the time required by the algorithms. Again, it is clear that the QBF-based algorithm outperforms the other two. We still observe that the median time is smaller than the average time, which indicates that significant variability still exists. Curiously, the times depend on the parity of the formula sizes: even-size formulae are easier than odd-size formulae. The reason could be that, except in the case of negation, the rest of the connectives are binary. 

\begin{figure}[t]
\hspace{-2mm}
\includegraphics[width=0.52\textwidth]{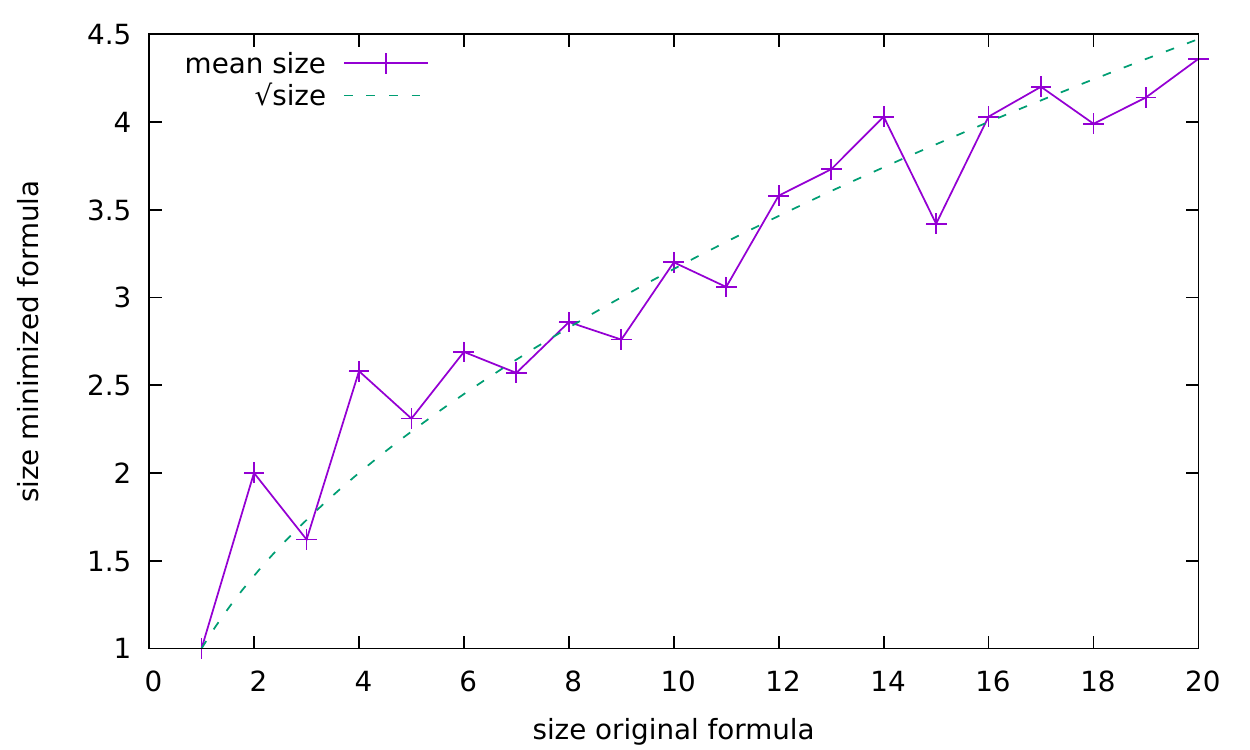}
\hspace{-20mm}\hfill
\includegraphics[width=0.52\textwidth]{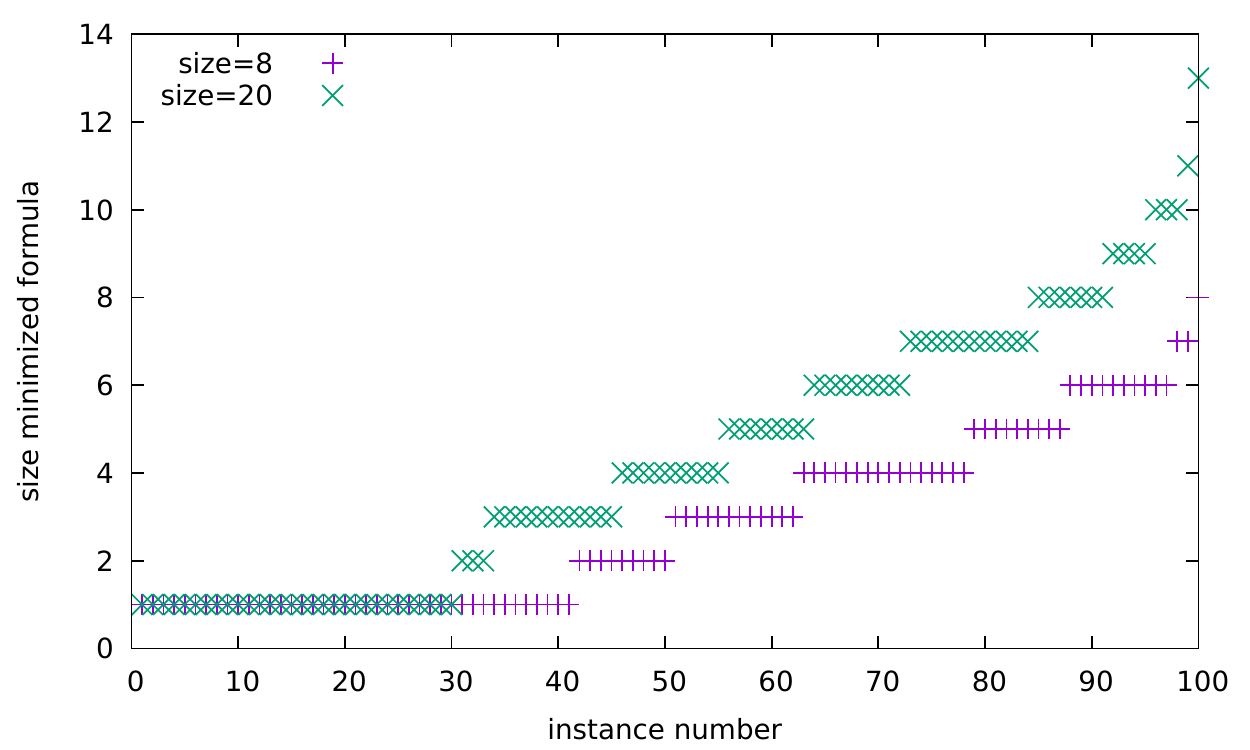}\hspace{-2mm}
\caption{Average size of the minimized formula w.r.t. the size of the original formula (left) and distribution of minimized sizes for formulae of original size $8$ and $20$ (right).}\label{fig:sizes}
\end{figure}

In Figure~\ref{fig:sizes} (left), we show how the average size of the minimized formula grows with respect to the size of the original formula. We observe that the growth is close to the square root of the original size. Recall that we generate random formulae of size $s$ and with $\sqrt{s}$ variables. Curiously, we observe that odd-size formulae are simplified more than even-size formulae, although the reason for this is not clear. In Figure~\ref{fig:sizes} (right), we show the distribution of sizes of the minimized formulae (for original formulae of sizes $20$ and $8$).

\begin{figure}[t]
\hspace{-2mm}
\includegraphics[width=0.52\textwidth]{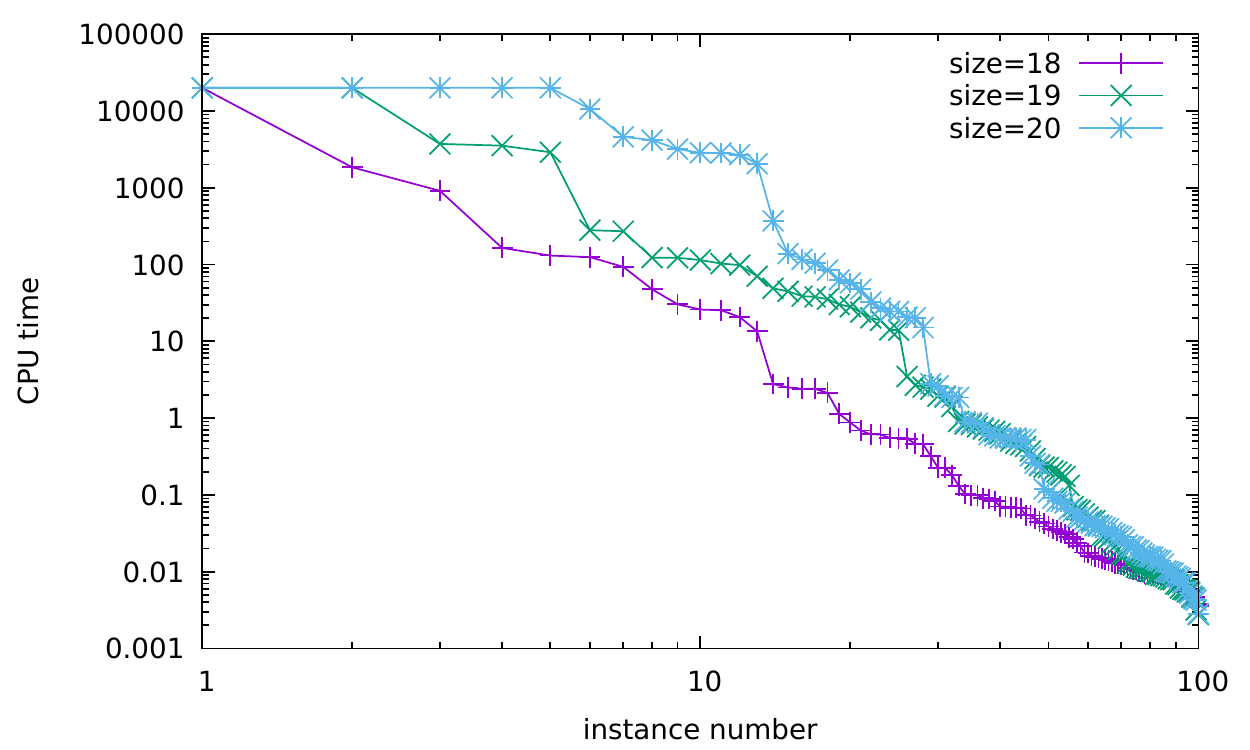}
\hspace{-20mm}\hfill
\includegraphics[width=0.52\textwidth]{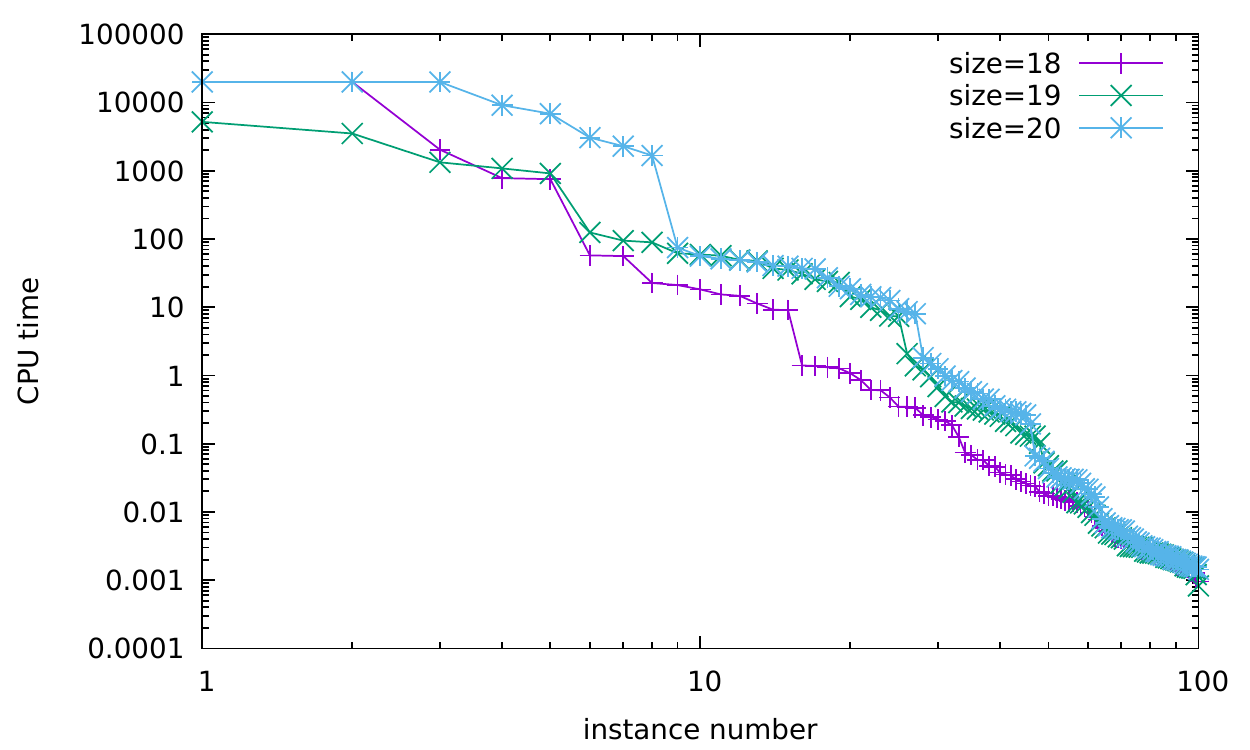}
\hspace{-4mm}
\caption{Distribution of times for the instances solved with the brute-force algorithm (left) and SAT-based algorithm (right).}\label{fig:disttimes}
\end{figure}

As mentioned above, we observe significant variability in CPU times, for the brute-force and SAT-based algorithms. In Figure~\ref{fig:disttimes}, we sort the instances in decreasing order of CPU time and represent, in double logarithmic axes, these times for the $100$ instances. We observe that this representation is close to a line (truncated on the top due to time-outs) with an increasing (negative) slope when the size increases. This implies that the CPU time in these algorithms follows a power-law probability distribution, where the time required by a few instances is responsible for most of the average time. The standard solution in these situations is to use some kind of restart policy or some randomization of the algorithm. In our case, we could randomize the order of the candidates to minimal formulae. However, since we want to obtain the minimal equivalent formula, we cannot randomize the order of the sizes of formulae that we try. 

\section{Conclusion and Future Work}\label{sec:conclusions}

In this paper, we have analyzed the practical use of three algorithms for general Boolean formula minimization. A simple algorithm that proves that the problem is in $\Sigma^p_2$, one based on the use of a SAT solver to check formula equivalences, and one that uses a Tseitin encoding of a formula's scheme and a QBF solver. We show that the third one clearly outperforms the other two. Therefore, the use of QBF solvers represents the state-of-the-art for the Boolean minimization problem.

Our experiments have been limited to Boolean formulae. The first natural extension of this work would be to see if this or similar methods could scale up to other (more expressive) formalisms, e.g., first-order logic (FOL). This would open up a range of interesting research questions, as in FOL, equivalence is undecidable. Adapting the QBF approach would probably not be feasible, yet, a semi-brute force approach, e.g., using a first-order theorem prover, could prove successful.

\bibliographystyle{splncs04}
\bibliography{cade23}

\end{document}